\documentclass[sigconf,nonacm]{acmart}
\AtBeginDocument{%
  \providecommand\BibTeX{{%
    \normalfont B\kern-0.5em{\scshape i\kern-0.25em b}\kern-0.8em\TeX}}}

\setcopyright{acmcopyright}
\copyrightyear{2023}
\acmYear{2023}

\acmConference[WSDAIF 2023]{2nd Workshop on Synthetic Data for AI in Finance}{Nov 27,
  2023}{Brooklyn, NY}
\begin{document}

\title{FiFAR: A Fraud Detection Dataset for Learning to Defer}

\author{Jean V. Alves}
\affiliation{%
  \institution{Feedzai}
  \institution{Instituto Superior Técnico}
  \city{Universidade de Lisboa}
  \country{Portugal}
}
\email{jean.alves@feedzai.com}

\author{Diogo Leitão}
\affiliation{%
  \institution{Feedzai}
  \city{Lisboa}
  \country{Portugal}
}
\email{diogo.leitao@feedzai.com}

\author{S\'ergio Jesus}
\affiliation{%
  \institution{Feedzai}
  \city{Lisboa}
  \country{Portugal}
}
\email{sergio.jesus@feedzai.com}

\author{Marco O. P. Sampaio}
\affiliation{%
  \institution{Feedzai}
  \city{Lisboa}
  \country{Portugal}
}
\email{marco.sampaio@feedzai.com}

\author{Pedro Saleiro}
\affiliation{%
  \institution{Feedzai}
  \city{Lisboa}
  \country{Portugal}
}
\email{pedro.saleiro@feedzai.com}

\author{Mário A. T. Figueiredo}
\affiliation{%
  \institution{Instituto Superior Técnico}
  \city{Universidade de Lisboa}
  \country{Portugal}
}

\email{mario.figueiredo@tecnico.ulisboa.pt}

\author{Pedro Bizarro}
\affiliation{%
  \institution{Feedzai}
  \city{Lisboa}
  \country{Portugal}
}
\email{pedro.bizarro@feedzai.com}

\renewcommand{\shortauthors}{Jean V. Alves, et al.}

\begin{abstract}
  Public dataset limitations have significantly hindered the development and benchmarking of \textit{learning to defer} (L2D) algorithms, which aim to optimally combine human and AI capabilities in hybrid decision-making systems. In such systems, human availability and domain-specific concerns introduce difficulties, while obtaining human predictions for training and evaluation is costly. Financial fraud detection is a high-stakes setting where algorithms and human experts often work in tandem; however, there are no publicly available datasets for L2D concerning this important application of human-AI teaming. To fill this gap in L2D research, we introduce the \textit{Financial Fraud Alert Review} Dataset (FiFAR), a synthetic bank account fraud detection dataset, containing the predictions of a team of 50 highly complex and varied synthetic fraud analysts, with varied bias and feature dependence. We also provide a realistic definition of human work capacity constraints, an aspect of L2D systems that is often overlooked, allowing for extensive testing of assignment systems under real-world conditions.
  We use our dataset to develop a capacity-aware L2D method and rejection learning approach under realistic data availability conditions, and benchmark these baselines under an array of 300 distinct testing scenarios. We believe that this dataset will serve as a pivotal instrument in facilitating a systematic, rigorous, reproducible, and transparent evaluation and comparison of L2D methods, thereby fostering the development of more synergistic human-AI collaboration in decision-making systems. The public dataset and detailed synthetic expert information are available at: \url{https://github.com/feedzai/fifar-dataset}
\end{abstract}


\maketitle

\section{Introduction}
Recently, an increasing body of research has been dedicated to studying human-AI collaboration (HAIC), with several authors arguing that humans have complementary sets of strengths and weaknesses to those of AI \cite{de-arteagaCaseHumansintheLoopDecisions2020, dellermannHybridIntelligence2019}. Collaborative systems have demonstrated that humans are able to rectify model predictions in specific instances \cite{de-arteagaCaseHumansintheLoopDecisions2020}, and have shown that humans, in collaboration with ML models, may achieve synergistic performance - a higher performance than the expert or the model on their own \cite{inkpen2022advancing}. 

The state-of-the-art approach to manage assignments in human-AI collaboration is \textit{learning to defer} (L2D)
\cite{charusaieSampleEfficientLearning2022, hemmerFormingEffectiveHumanAI2022, raghu2019directuncertaintyprediction, raghuAlgorithmicAutomationProblem2019, mozannarConsistentEstimatorsLearning2020, mozannar2023should, madrasPredictResponsiblyImproving2018}. These are algorithms that choose whether to assign an instance to a human or a ML model, aiming to take advantage of their complementary strengths. L2D algorithms require large amounts of data on human decisions: some require multiple human predictions per instance \cite{raghu2019directuncertaintyprediction, raghuAlgorithmicAutomationProblem2019}, while others often require human predictions to exist for every single training instance \cite{madrasPredictResponsiblyImproving2018, mozannarConsistentEstimatorsLearning2020, vermaCalibratedLearningDefer2022, hemmerFormingEffectiveHumanAI2022}. Due to the unavailability of large datasets containing human predictions, and the cost of obtaining large amounts of data annotated by human experts, these methods are frequently developed with small datasets, containing limited human predictions, or by using synthetic human subjects. The synthesized expert behavior is often simplistic, and varies significantly between authors. Consequently, research into L2D is lacking in robust benchmarking of different methods. 

Financial fraud detection is a high-stakes use case where human-AI collaboration is often applied. Machine Learning models can be used in anti-money laundering, where an automated system monitors transactions, raising alerts that are then reviewed by human-experts \cite{kute2021deep}. In e-commerce transaction fraud detection, ML models' advice may help improve the accuracy of human decision-makers, as well as expedite the decision making process \cite{amarasinghe2022importanceofapplication}. However, research into applying L2D in fraud prevention settings is lacking, possibly due to a lack of adequate public datasets in this domain.

To address this issue, we present the \textit{Financial Fraud Alert Review} (FiFAR) Dataset, which includes the predictions of a team of 50 highly complex synthetic fraud analysts, generated in order to simulate a wide variety of human behaviours. We use a novel approach to generate complex synthetic experts, with control over performance, feature dependence and bias towards a protected attribute; and define capacity constraints limiting the amount of instances that can be deferred to each expert. We also create a version of our dataset simulating realistic data availability conditions (only one expert prediction per instance) during training, thus providing realistic training and testing scenarios for L2D research. Subject to these conditions, we develop a capacity aware L2D algorithm, and benchmark two versions of our method as well as a capacity aware version of \textit{rejection learning}, by testing their performance and fairness under 300 different testing scenarios. We hope to bolster research into development and testing of L2D methods subject to real-world problems, such as changes in human availability and limited amounts of human prediction data. Our dataset is available at: \url{https://github.com/feedzai/fifar-dataset}

\section{Background and Related Work}

In this section we discuss the most commonly used datasets in HAIC research, methods of synthetic expert generation in L2D research, and the current state-of-the-art L2D approaches.

\subsection{Current HAIC Datasets}

A Dataset suitable for L2D training and evaluation has to comply with a few requirements. Firstly, the dataset must contain a sizeable amount of predictions from each member of the expert team, to enable modeling of the human behavior. The human which made each prediction must be identifiable, allowing for individual modelling of each expert's behavior. Finally, in testing, we must have a set of each expert's predictions for every instance in the test set, as the assignment system may query any expert on a given instance. To the best of our knowledge, there are only two public real-world datasets suitable for training multi-expert human-AI assignment systems, which we now describe. 
The NIH Clinical Center X-ray dataset \cite{wangChestxray8HospitalscaleChest2017}, used by Hemmer et al. \cite{hemmerFormingEffectiveHumanAI2022}, is a computer vision dataset aimed at detecting airspace opacity. For each X-ray image, there are recorded predictions from an ensemble of 22 experts, and a golden label created by an independent team of 3 radiologists.
The main drawback of this dataset is its size: it contains only 4,374 X-ray images. 
The Hate Speech and Offensive Language Detection dataset enriched by Keswani et al. \cite{keswaniUnbiasedAccurateDeferral2021}, consists of a subset of 1,471 tweets from
the original dataset \cite{davidsonAutomatedHateSpeech2017}, annotated by a total of 170 crowd-sourcing workers according to the presence of hate speech or offensive language. Each tweet was labelled by an average of 10 workers, meaning that each worker labelled an average of 87 instances. The low volume of instances hinders the capacity to model individual expertise conditioned to the input space.


\subsection{Simulation of Human Experts}

Due to the lack of adequate real-world datasets, several authors have resorted to synthesizing expert behavior for datasets in the ML literature. Madras et al. \cite{madrasPredictResponsiblyImproving2018} propose a \textit{model-as-expert} technique, fitting a ML classifier to mimic expert behavior on two binary classification datasets (COMPAS \cite{kirchnerHowWeAnalyzed2017} and Heritage Health \cite{HeritageHealthPrize}). They use the same ML algorithm used for the main task with extra features, in order to simulate access to exclusive information. These authors also introduce bias, with the goal of studying unfairness mitigation. Verma and Nalisnick \cite{vermaCalibratedLearningDefer2022} use a similar approach to produce an expert on the HAM10000 dataset \cite{tschandlHAM10000DatasetLarge2018}. In a \textit{model-as-expert} approach, the modelling bias of the ML algorithm is the same for the main classifier and the synthetic experts. Furthermore, they are trained on data with a large fraction of features in common. This may lead to artificially large agreement between the classifier and experts, which is why we choose not to use it. 

Other authors use a \textit{label noise} approach to produce arbitrarily accurate expert predictions. Mozannar and Sontag use CIFAR-10
\cite{krizhevskyLearningMultipleLayers2009}, where they simulate an expert with perfect accuracy on a fraction of the 10 classes, but random accuracy on the others (see also Verma and Nalisnick \cite{vermaCalibratedLearningDefer2022} and Charusaie et al. \cite{charusaieSampleEfficientLearning2022}). 
The main drawback of these synthetic experts is that their expertise is either feature-independent or only dependent on a single feature or concept. As such, the methods tested on these benchmarks are not being challenged to learn nuanced and varied types of expertise.
This type of approach
has been criticised. Zhu et al. \cite{zhu2021second} and Berthon et al. \cite{berthon2021confidence} argue that \textit{instance-dependent label noise} (IDN) is more realistic, as human error is more likely to be dependent on the difficulty of a given task, and, as such, should also be dependent on the instance's features. Our approach will make use of \textit{instance-dependent label noise}.

\subsection{Current L2D Methods}

One of the simplest deferral approaches in the literature is given by \textit{rejection learning} (ReL) \cite{RL_original_chowOptimumRecognitionError1970, RL_cortesLearningRejection2016}. 
In a human-AI collaboration setting, ReL defers instances from the model to humans \cite{madrasPredictResponsiblyImproving2018, raghuAlgorithmicAutomationProblem2019}. 
Its simplest implementation \cite{RL_hendrycksBaselineDetectingMisclassified2017} produces uncertainty estimates for the model's prediction in each instance, ranking them, and rejecting to predict if the uncertainty is above a given threshold \cite{RL_original_chowOptimumRecognitionError1970, RL_cortesLearningRejection2016}.

\begin{figure*}
\centering
  \includegraphics[width=0.85\textwidth]{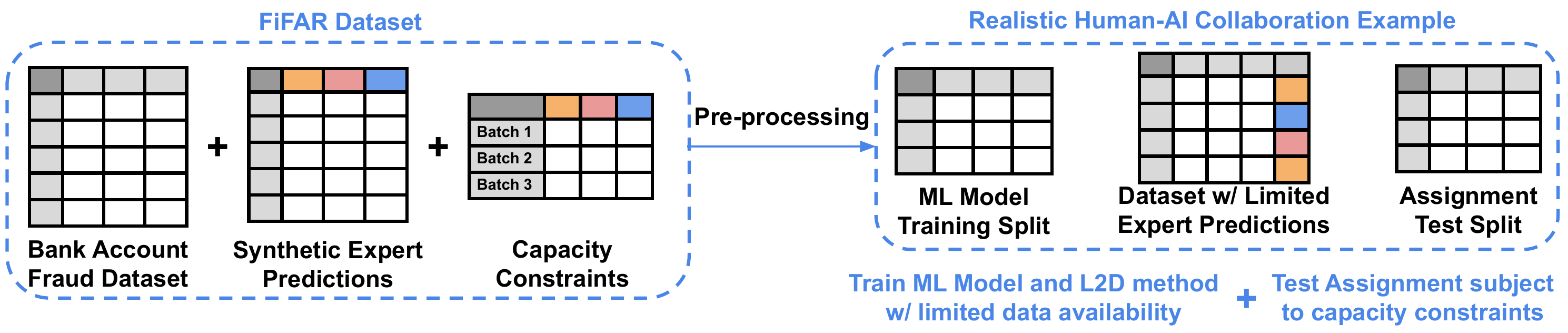}
  \caption{FiFAR Dataset}
  \Description{.}
  \label{fig:teaser}
\end{figure*}

Madras et al. \cite{madrasPredictResponsiblyImproving2018} argue that ReL is sub-optimal because it does not consider the performance of the human(s) involved in the task, so they propose \textit{learning to defer} (L2D). In the original L2D framework, the classifier and assignment system are jointly trained, taking into account a single model and a single human. Many authors have since contributed to the single-expert framework \cite{mozannarConsistentEstimatorsLearning2020, vermaCalibratedLearningDefer2022}. Keswani et al. \cite{keswaniUnbiasedAccurateDeferral2021} observe that decisions can often be deferred to one or more humans out of a team, expanding L2D to the multi-expert setting \cite{keswaniUnbiasedAccurateDeferral2021}, followed by \citet{hemmerFormingEffectiveHumanAI2022, verma2023learning}.
Most L2D approaches require predictions from every human team member, for every training instance, imposing significant, and often unrealistic, data requirements \cite{leitao2022human}. Furthermore, the limited work capacity of each team member often goes unaddressed \cite{leitao2022human}.

In conclusion currently available L2D algorithms require realistic datasets for development and testing under realistic conditions. As it is currently unfeasible to collect real world expert predictions for large datasets, a promising avenue is to develop methods to generate synthetic expert predictions that look realistic.

\section{Dataset and HAIC Scenario}

The FiFAR Dataset is comprised by three components, represented in Figure \ref{fig:teaser}: a base dataset, which contains each instances features; a table containing each synthetic expert's predictions for each instance (see Section \ref{section: simexp}); and a set of capacity constraints, detailing the amount of instances that each expert can process in a given time interval (see Section \ref{section: capconstr}). Using our dataset, we instantiate a scenario simulating the development and testing of a L2D system under realistic conditions (see Section \ref{section: haic setup}). 

\subsection{Base Dataset}
\label{section: Base Dataset}
As the base dataset, we choose to use the publicly available bank-account-fraud tabular dataset \citep{jesusTurningTablesBiased2022}. This dataset is sizeable (one million rows) and boasts other key properties that are relevant for our use case. 
The data was generated by applying tabular data generation techniques on an anonymized, real-world bank account opening fraud detection dataset.
Each instance represents a bank account opening application, with features containing information about the application and the applicant, and  a label that denotes if the instance is a fraudulent (1) or a legitimate (0) application. 

The task of a decision maker (automated or human) is to either accept (predicted negative) or reject (predicted positive) it. A positive prediction results in a declined application. As such, false positives in account opening fraud can significantly affect a person’s life (with no possibility to open a bank account or to access credit). This is thus a cost-sensitive problem, where the cost of a false positive must be weighed against the cost of a false negative. The optimization goal is to maximize recall at a fixed FPR (we use 5\%), which implicitly establishes a relationship between the costs. This task also entails fairness concerns, as ML models trained on this dataset tend to raise more false fraud alerts for older clients ($\geq 50$ years), thus reducing their access to a bank account.

\subsection{Decision Generation Method}
\label{section: simexp}

Our expert generation approach is based on \textit{instance-dependent noise}, in order to obtain more realistic experts, whose probability of error varies with the properties of each instance.
We generate synthetic predictions by flipping each label $y_i$ with probability $\mathbb{P} (m_{i,j}\neq y_i \lvert \pmb{x}_i, y_i)$. In some HAIC systems, the model score for a given instance may also be shown to the expert \cite{amarasinghe2022importanceofapplication, de-arteagaCaseHumansintheLoopDecisions2020, levy2021assessing}, so an expert's decision may also be dependent on an ML model score $m(\pmb{x}_i)$. We define the expert's probabilities of error, for a given instance, as a function of its features, $\pmb{x}_i$, and the model score $m(\pmb{x}_i)$,
\begin{equation*}
    \begin{cases}
    \mathbb{P}(m_{i,j} = 1 \lvert y_i = 0, \pmb{x}_i, M) = \sigma \Big( \beta_{0} - \alpha  \frac{\pmb{w} \cdot \pmb{x}_i + w_{M}M(\pmb{x} _ i)}{\sqrt{\pmb{w} \cdot \pmb{w}  + w_{M}^2}}   \Big )\\
    
    \mathbb{P}(m_{i,j} = 0 \lvert y_i = 1, \pmb{x}_i, M) = \sigma \Big( \beta_{1} + \alpha  \frac{\pmb{w} \cdot \pmb{x}_i + w_{M}M(\mathbf{\pmb{x}}_i)}{\sqrt{\pmb{w} \cdot \pmb{w}  + w_{M}^2}}  \Big ) ,
    
    \end{cases} 
\end{equation*}
\begin{equation*}
        M(\mathbf{x}_i) = \begin{cases}
        \frac{m(\pmb{x}_i) - t}{2t} \;, \quad m \leq t \vspace{1mm}\\
        \frac{m(\pmb{x}_i) - t}{2(1-t)}, \quad m > t,
    \end{cases} \; 
\end{equation*}
where $\sigma$ denotes a sigmoid function and $M$ is a transformed version of the original model score $m \in [0,1]$. Each expert's probabilities of the two types of error are parameterized by five variables: $\beta_0, \beta_1,\alpha,\mathbf{w}$ and $w_M$. The weight vector $\mathbf{w}$ 
embodies a relation between the features and the probability of error. To impose a dependence on the model score, we can set $w_{M} \neq 0$. The feature weights are normalized so that we can separately control, with the $\alpha$ parameter, the overall magnitude of the variation of the probability of error due to the instance's features. The values of $\beta_1$ and $\beta_0$ control the base probability of error. 

\subsection{Human Decision-Making Properties}

In this section we list the characteristics of human decision-making that we aim to capture with our approach, in order to make our synthetic experts as realistic as possible.

\textbf{Feature and AI assistant dependence}:
When a decision is made by an expert, it is assumed that they will base themselves on information related to the instance in question. Therefore, we expect experts to be dependent on the instance's features.

In some real world deferral systems \citep{amarasinghe2022importanceofapplication, de-arteagaCaseHumansintheLoopDecisions2020}, the instance's features are accompanied by an AI model's score, representing the model's estimate of the probability that said instance belongs to the positive class. The aim of presenting the model score to an expert is to provide them with extra information, as well as possibly expediting the decision process. It has been shown that, in this scenario, expert's performance can be impacted by presenting the model's score when deferring a case to an expert \citep{amarasinghe2022importanceofapplication, de-arteagaCaseHumansintheLoopDecisions2020, levy2021assessing}. We should then consider that our experts are impacted by the ML classifier.

\textbf{AI assistance and algorithmic bias}:
Should the generated experts use an AI assistant, we expect experts not to be in perfect agreement with the model, due to the assumption that humans and models have complementary strengths and weaknesses \citep{de-arteagaCaseHumansintheLoopDecisions2020, dellermannHybridIntelligence2019}. As such, we would assume humans and AI perform better than one another in separate regions of the feature space.
The degree of "algorithmic bias" \citep{alon2023humanaiinteractions} varies between humans, measured as the model's impact on a human's performance \citep{jacobs2021machine, inkpen2022advancing}. As such, our synthetic experts also exhibit varying levels of model dependence.

\textbf{Varied Expert Performance}
In order for our team of experts to be realistic, it is important that these exhibit varying levels of overall performance. Experts within a field have been shown to have varying degrees of expertise, with some being outperformed by ML models \citep{goel2021accuracy, gulshan2016development}.
As such human decision processes can be expected to be varied even amongst a team of experts.

\textbf{Predictability and Consistency}
It is a common assumption that, when making a decision, experts follow an internal process based on the available information. However, it is also known that even experts are still subject to flaws that are inherent to human decision making processes, one of these being inconsistency. When presented with similar cases, at different times, experts may perform drastically different decisions \citep{hungryjudge, grimstad2007inconsistency}. Therefore we can expect a human's decision making process not to be entirely deterministic.

\textbf{Human Bias and Unfairness}
It is also important to consider the role that the assignment system can play in mitigating unfairness. If an expert can be determined to be particularly unfair with respect to a given protected attribute, the assignment system can learn not to defer certain cases to that expert. In order to test the fairness of the system as a whole, it is useful to create a team of individuals with varying propensity for unfair decisions.

To simulate a wide variety of human behavior, we created four distinct expert groups. All groups have similar performance as measured by their TPR and FPR, with a fraction of the team performing worse than the ML Model. The first is a \textit{Standard} expert group: on average as unfair as the model, dependent on model score and, on average, twelve different features. The three other types of expert are variations on the \textit{Standard} group. i) Unfair: experts which are more likely to incorrectly reject an older customer's application. ii) \textit{Model agreeing}: experts which are heavily swayed by the model score. iii) \textit{Sparse}: experts which are dependent on fewer features.

\subsection{Definition of Capacity Constraints}
\label{section: capconstr}
Firstly, we formalize how to define capacity constraints. Humans are limited in the number of instances they may process in any given time period
(e.g., work day). In real-world systems, human capacity must be applied over batches of instances, not over the whole dataset at once (e.g. balancing the human workload over an entire month is not the same as balancing it daily). A real-world assignment system must then process instances taking into account the human limitations over a given “batch” of cases, corresponding to a pre-defined time period. 
We divide our dataset into several batches and, for each batch, define the maximum number of instances that can be processed by each expert. In any given dataset comprised of $N$ instances, capacity constraints can be represented by a vector $ \pmb{b}$, where component $b_i$ denotes which batch instance $i \in \{1,...,N\}$ belongs to, as well as a human capacity matrix $H$, where element $H_{b,j}$ is a non-negative integer denoting the number of instances expert $j$ can process in batch $b$.

To define the batch vector, we have to define the number of cases in each batch, then distribute instances throughout the batches. To define the capacity matrix, we consider 4 separate parameters. (1) Deferral\_rate: maximum fraction of each batch that can be deferred to the human team; (2) Distribution \textit{homogeneous} or \textit{variable}. Should the distribution be \textit{homogeneous}, every expert will have the same capacity; otherwise, each expert's capacity is sampled from $\mathcal{N}(\mu_{d} = \text{Deferral\_rate} \times \text{Batch\_Size} / N_{experts}$,  $0.2 \times \mu_{d})$, chosen so that each expert's capacity fluctuates around the value corresponding to an homogeneous distribution; (3) Absence rate, defined as the fraction of experts that are absent in each batch. This allows for testing several different team configurations without generating new experts, or scenarios where not all experts work in the same time periods. (4) Expert Pool, defined as which types of experts (standard, unfair, sparse or model agreeing) can be part of the team.

To allow for extensive testing, we create a vast set of capacity constraints. In Table ~\ref{table: baseline results}, under ``Scenario Properties", we list the different combinations of settings used. For each combination, several seeds were set for the batch, expert absence, and capacity sampling.

\begin{table*}
  \centering
  \caption{Baseline Results. Intervals denote standard deviation. FPR disparity ($\mbox{FPR}_d$) standard deviations are omitted due to low variability. ``Model Only'' represents a fully automated baseline, with predictions made by the model, according to threshold $t$.}
  \label{table: baseline results}
  \makebox[\textwidth][c]{
  \begin{tabular}{ccccccccccccc}
  
    \toprule
     \multicolumn{5}{c}{Scenario Properties}& \multicolumn{2}{c}{Model Only} & \multicolumn{2}{c}{$\mbox{ReL}$} &\multicolumn{2}{c}{$\mbox{ReL}_{\mbox{\scriptsize greedy}}$} & \multicolumn{2}{c}{$\mbox{ReL}_{\mbox{\scriptsize linear}}$} \\
    \cmidrule(r){1-5}
    \cmidrule(r){6-7}
    \cmidrule(r){8-9}
    \cmidrule(r){10-11}
    \cmidrule(r){12-13}
    Pool& Batch Size & Deferral Rate & Absence Rate & $\sigma_{d}$ & Loss & $\mbox{PE}$ & Loss & $\mbox{PE}$ & Loss & $\mbox{PE}$ &Loss & $\mbox{PE}$ \\
    \midrule
      all & 250 & 0.2 & 0.0 & 0.2 & 918 & 0.33 & $\mathbf{753}\mbox{\scriptsize{$\pm12$} }$ & $0.29$ & $780\mbox{\scriptsize{$\pm10$} }$ & $0.31$ & $780\mbox{\scriptsize{$\pm9$} }$ & $0.32$ \\ 
all & 250 & 0.2 & 0.0 &0.0& 918 & 0.33 & $\mathbf{755}\mbox{\scriptsize{$\pm14$} }$ & $0.30$ & $781\mbox{\scriptsize{$\pm8$} }$ & $0.31$ & $789\mbox{\scriptsize{$\pm8$} }$ & $0.32$ \\ 
all & 250 & 0.2 & 0.5 & 0.2 & 918 & 0.33 & $\mathbf{760}\mbox{\scriptsize{$\pm11$} }$ & $0.29$ & $788\mbox{\scriptsize{$\pm9$} }$ & $0.31$ & $782\mbox{\scriptsize{$\pm9$} }$ & $0.32$ \\ 
all & 250 & 0.2 & 0.5 &0.0& 918 & 0.33 & $\mathbf{768}\mbox{\scriptsize{$\pm10$} }$ & $0.29$ & $788\mbox{\scriptsize{$\pm11$} }$ & $0.31$ & $786\mbox{\scriptsize{$\pm10$} }$ & $0.32$ \\ 
all & 250 & 0.5 & 0.0 & 0.2 & 918 & 0.33 & $\mathbf{746}\mbox{\scriptsize{$\pm17$} }$ & $0.29$ & $788\mbox{\scriptsize{$\pm7$} }$ & $0.34$ & $766\mbox{\scriptsize{$\pm9$} }$ & $0.36$ \\ 
all & 250 & 0.5 & 0.0 &0.0& 918 & 0.33 & $\mathbf{759}\mbox{\scriptsize{$\pm14$} }$ & $0.29$ & $790\mbox{\scriptsize{$\pm4$} }$ & $0.34$ & $765\mbox{\scriptsize{$\pm13$} }$ & $0.36$ \\ 
all & 250 & 0.5 & 0.5 & 0.2 & 918 & 0.33 & $\mathbf{756}\mbox{\scriptsize{$\pm13$} }$ & $0.29$ & $779\mbox{\scriptsize{$\pm7$} }$ & $0.32$ & $782\mbox{\scriptsize{$\pm8$} }$ & $0.36$ \\ 
all & 250 & 0.5 & 0.5 &0.0& 918 & 0.33 & $\mathbf{754}\mbox{\scriptsize{$\pm11$} }$ & $0.29$ & $783\mbox{\scriptsize{$\pm6$} }$ & $0.32$ & $783\mbox{\scriptsize{$\pm5$} }$ & $0.36$ \\ 
all & 5000 & 0.2 & 0.0 & 0.2 & 918 & 0.33 & $\mathbf{752}\mbox{\scriptsize{$\pm8$} }$ & $0.30$ & $780\mbox{\scriptsize{$\pm4$} }$ & $0.32$ & $779\mbox{\scriptsize{$\pm5$} }$ & $0.33$ \\ 
all & 5000 & 0.2 & 0.0 &0.0& 918 & 0.33 & $\mathbf{752}\mbox{\scriptsize{$\pm12$} }$ & $0.30$ & $778\mbox{\scriptsize{$\pm3$} }$ & $0.32$ & $782\mbox{\scriptsize{$\pm5$} }$ & $0.33$ \\ 
all & 5000 & 0.2 & 0.5 & 0.2 & 918 & 0.33 & $\mathbf{762}\mbox{\scriptsize{$\pm10$} }$ & $0.30$ & $778\mbox{\scriptsize{$\pm12$} }$ & $0.31$ & $773\mbox{\scriptsize{$\pm4$} }$ & $0.33$ \\ 
all & 5000 & 0.2 & 0.5 &0.0& 918 & 0.33 & $\mathbf{753}\mbox{\scriptsize{$\pm9$} }$ & $0.30$ & $776\mbox{\scriptsize{$\pm11$} }$ & $0.31$ & $776\mbox{\scriptsize{$\pm3$} }$ & $0.33$ \\ 
all & 5000 & 0.5 & 0.0 & 0.2 & 918 & 0.33 & $\mathbf{749}\mbox{\scriptsize{$\pm8$} }$ & $0.29$ & $774\mbox{\scriptsize{$\pm6$} }$ & $0.34$ & $768\mbox{\scriptsize{$\pm6$} }$ & $0.36$ \\ 
all & 5000 & 0.5 & 0.0 &0.0& 918 & 0.33 & $\mathbf{750}\mbox{\scriptsize{$\pm11$} }$ & $0.29$ & $776\mbox{\scriptsize{$\pm8$} }$ & $0.34$ & $768\mbox{\scriptsize{$\pm1$} }$ & $0.36$ \\ 
all & 5000 & 0.5 & 0.5 & 0.2 & 918 & 0.33 & $\mathbf{759}\mbox{\scriptsize{$\pm12$} }$ & $0.29$ & $774\mbox{\scriptsize{$\pm7$} }$ & $0.32$ & $780\mbox{\scriptsize{$\pm8$} }$ & $0.37$ \\ 
all & 5000 & 0.5 & 0.5 &0.0& 918 & 0.33 & $\mathbf{758}\mbox{\scriptsize{$\pm8$} }$ & $0.29$ & $773\mbox{\scriptsize{$\pm6$} }$ & $0.33$ & $781\mbox{\scriptsize{$\pm7$} }$ & $0.37$ \\ 
    
    \bottomrule
  \end{tabular}
  }
  \newline

\end{table*}

\begin{table*}
  
  \centering
  \caption{Varying Expert Pool Results. Nomenclature used is consistent with Table \ref{table: baseline results}}
  \label{table: baseline results pool}
  \makebox[\textwidth][c]{
  \begin{tabular}{ccccccccccccc}
  
    \toprule
     \multicolumn{5}{c}{Scenario Properties}& \multicolumn{2}{c}{$\mbox{ReL}$} &\multicolumn{2}{c}{$\mbox{ReL}_{\mbox{\scriptsize greedy}}$} & \multicolumn{2}{c}{$\mbox{ReL}_{\mbox{\scriptsize linear}}$} \\
    \cmidrule(r){1-5}
    \cmidrule(r){6-7}
    \cmidrule(r){8-9}
    \cmidrule(r){10-11}
    \cmidrule(r){12-13}
    Pool& Batch Size & Deferral Rate & Absence Rate & $\sigma_{d}$   & Loss & $\mbox{PE}$ & Loss & $\mbox{PE}$ &Loss & $\mbox{PE}$ \\
    \midrule
    agreeing & 250 & 0.2 & 0.0 &0.0& $813\mbox{\scriptsize{$\pm8$} }$ & $0.37$ & $873\mbox{\scriptsize{$\pm7$} }$ & $0.37$ & $\mathbf{810}\mbox{\scriptsize{$\pm3$} }$ & $0.34$ \\ 
agreeing & 250 & 0.5 & 0.0 &0.0& $815\mbox{\scriptsize{$\pm11$} }$ & $0.39$ & $900\mbox{\scriptsize{$\pm4$} }$ & $0.40$ & $\mathbf{783}\mbox{\scriptsize{$\pm5$} }$ & $0.36$ \\ 
agreeing & 5000 & 0.2 & 0.0 &0.0& $816\mbox{\scriptsize{$\pm7$} }$ & $0.37$ & $875\mbox{\scriptsize{$\pm4$} }$ & $0.37$ & $\mathbf{808}\mbox{\scriptsize{$\pm5$} }$ & $0.34$ \\ 
agreeing & 5000 & 0.5 & 0.0 &0.0& $814\mbox{\scriptsize{$\pm12$} }$ & $0.39$ & $905\mbox{\scriptsize{$\pm3$} }$ & $0.40$ & $\mathbf{784}\mbox{\scriptsize{$\pm3$} }$ & $0.36$ \\ 
sparse & 250 & 0.2 & 0.0 &0.0& $766\mbox{\scriptsize{$\pm9$} }$ & $0.29$ & $770\mbox{\scriptsize{$\pm6$} }$ & $0.31$ & $\mathbf{755}\mbox{\scriptsize{$\pm6$} }$ & $0.31$ \\ 
sparse & 250 & 0.5 & 0.0 &0.0& $752\mbox{\scriptsize{$\pm11$} }$ & $0.28$ & $738\mbox{\scriptsize{$\pm8$} }$ & $0.31$ & $\mathbf{737}\mbox{\scriptsize{$\pm11$} }$ & $0.34$ \\ 
sparse & 5000 & 0.2 & 0.0 &0.0& $752\mbox{\scriptsize{$\pm4$} }$ & $0.29$ & $767\mbox{\scriptsize{$\pm5$} }$ & $0.31$ & $\mathbf{738}\mbox{\scriptsize{$\pm3$} }$ & $0.32$ \\ 
sparse & 5000 & 0.5 & 0.0 &0.0& $764\mbox{\scriptsize{$\pm11$} }$ & $0.29$ & $758\mbox{\scriptsize{$\pm4$} }$ & $0.32$ & $\mathbf{737}\mbox{\scriptsize{$\pm5$} }$ & $0.34$ \\ 
standard & 250 & 0.2 & 0.0 &0.0& $\mathbf{742}\mbox{\scriptsize{$\pm13$} }$ & $0.30$ & $782\mbox{\scriptsize{$\pm12$} }$ & $0.32$ & $788\mbox{\scriptsize{$\pm7$} }$ & $0.33$ \\ 
standard & 250 & 0.5 & 0.0 &0.0& $\mathbf{739}\mbox{\scriptsize{$\pm9$} }$ & $0.31$ & $773\mbox{\scriptsize{$\pm9$} }$ & $0.33$ & $782\mbox{\scriptsize{$\pm6$} }$ & $0.34$ \\ 
standard & 5000 & 0.2 & 0.0 &0.0& $\mathbf{739}\mbox{\scriptsize{$\pm6$} }$ & $0.31$ & $773\mbox{\scriptsize{$\pm1$} }$ & $0.32$ & $773\mbox{\scriptsize{$\pm4$} }$ & $0.33$ \\ 
standard & 5000 & 0.5 & 0.0 &0.0& $\mathbf{731}\mbox{\scriptsize{$\pm12$} }$ & $0.31$ & $757\mbox{\scriptsize{$\pm2$} }$ & $0.33$ & $777\mbox{\scriptsize{$\pm4$} }$ & $0.35$ \\ 
unfair & 250 & 0.2 & 0.0 &0.0& $736\mbox{\scriptsize{$\pm8$} }$ & $0.22$ & $721\mbox{\scriptsize{$\pm6$} }$ & $0.24$ & $\mathbf{714}\mbox{\scriptsize{$\pm1$} }$ & $0.25$ \\ 
unfair & 250 & 0.5 & 0.0 &0.0& $722\mbox{\scriptsize{$\pm9$} }$ & $0.19$ & $708\mbox{\scriptsize{$\pm3$} }$ & $0.21$ & $\mathbf{687}\mbox{\scriptsize{$\pm8$} }$ & $0.23$ \\ 
unfair & 5000 & 0.2 & 0.0 &0.0& $724\mbox{\scriptsize{$\pm11$} }$ & $0.23$ & $726\mbox{\scriptsize{$\pm2$} }$ & $0.25$ & $\mathbf{711}\mbox{\scriptsize{$\pm2$} }$ & $0.26$ \\ 
unfair & 5000 & 0.5 & 0.0 &0.0& $712\mbox{\scriptsize{$\pm7$} }$ & $0.20$ & $712\mbox{\scriptsize{$\pm3$} }$ & $0.22$ & $\mathbf{682}\mbox{\scriptsize{$\pm5$} }$ & $0.24$ \\ 
    
    \bottomrule
  \end{tabular}
  }
\end{table*}

\subsection{HAIC Setup}
\label{section: haic setup}
\begin{figure}[t]
    \centering
    \includegraphics[width = 0.40\textwidth]{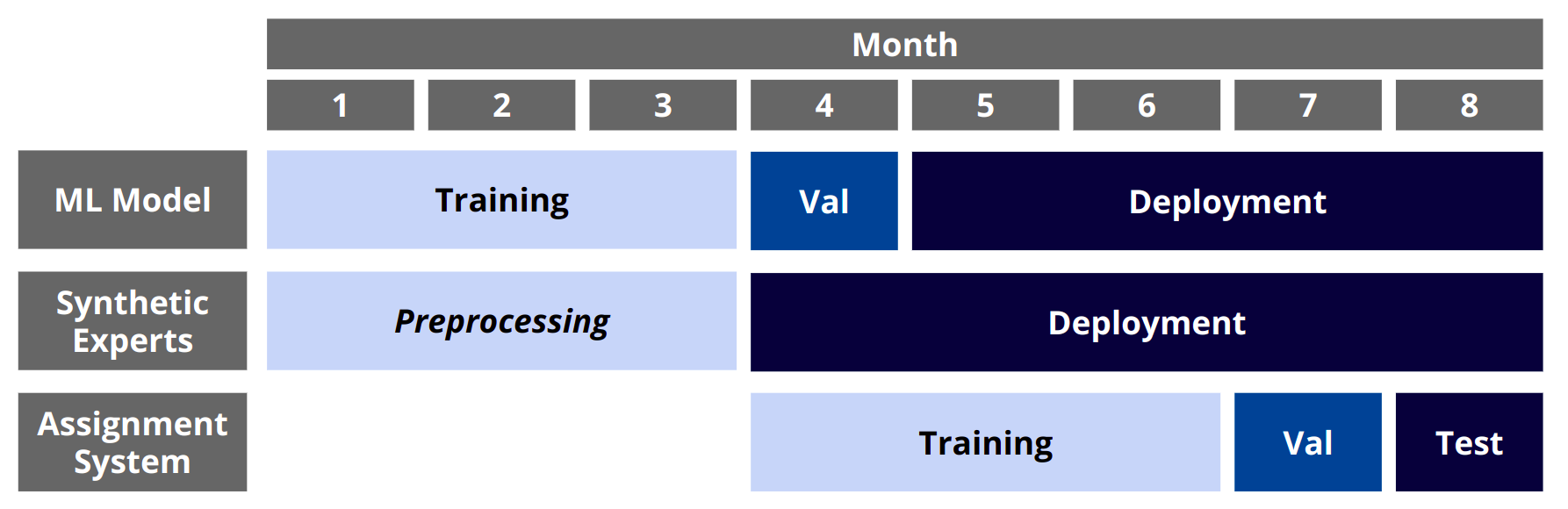}
    \caption{Temporal Splits for L2D Development}
    \label{fig:timeline}
\end{figure} 

We choose temporal splits of the dataset that aim to emulate a realistic scenario as close as possible, represented in figure \ref{fig:timeline}. To do so, we first train a fraud detection ML classifier. This model is trained on the first three months of the dataset and validated on the fourth month. We utilize the LightGBM \citep{keLightGBMHighlyEfficient2017} algorithm, due to its proven high performance on tabular data \citep{shwartz2022tabular, borisov2022deep}. The ML model yields a recall of 57.9\% in validation, for a threshold of $t = 0.051$, chosen in validation to obtain 5\% FPR. 

Our simulated experts are assumed to act alongside the ML model on the period ranging from the fourth to the eighth month.
There are several possible ways for models and humans to cooperate. In L2D testing, it is often assumed any instance can be deferred to either the model or the expert team. However, in a real world setting, it is common to use an ML model to raise alerts that are then reviewed by human experts \citep{de-arteagaCaseHumansintheLoopDecisions2020, han2020artificial}. Without an assignment method, the decision system would function as follows: a batch of instances is processed by the model, a fraction of the highest scoring instances are flagged for human review, and, finally, these instances are randomly assigned to experts, who make the final decision. The rest of the instances are automatically accepted. 

By assuming that the alert review system is employed from months four to seven, we can construct a dataset that would correspond to human predictions gathered over this period. Using this data, we train our assignment algorithms with the data of months four to six, validating them on the seventh month. Testing is done by creating a new deferral system, where the cases flagged by the ML classifier for review are distributed to the humans according to an assignment algorithm trained on the gathered data.

\section{Experimental Setting}

\subsection{Evaluating Assignments}

As stated in Section \ref{section: Base Dataset}, the optimization goal is to maximize the recall at a 5\% FPR (Neyman-Pearson Criterion). When evaluating a set of assignments, values for the FPR may not be the same across experiments. This hinders our ability to directly compare the recall of different algorithms (\textit{i.e.} If two methods obtain the same recall, the one with the lowest FPR is preferred). Implicitly, this optimization goal expresses a tradeoff between the misclassification costs of FP and FN mistakes, that is, a cost-sensitive problem. To evaluate performance, we can utilize a cost sensitive loss:
\begin{equation}
\label{eq: cost sensitive loss}
     L = \lambda \mbox{N}(\mbox{FP}) + \mbox{N}(\mbox{FN}) \quad \mbox{with} \quad \lambda = \frac{t}{1-t}\; ,
\end{equation}
Where N(FP/FN) is the number of FP/FN errors. The parameter $\lambda$ enforces a relationship between the cost of a FP and the cost of a FN. We now must define the relationship between our Neyman-Pearson criterion and the value of $\lambda$. Elkan \cite{elkanFoundationsCostSensitiveLearning2001} shows that 
the value of the ideal threshold $t$ for a binary classifier and the misclassification costs are related according to Equation ~\ref{eq: cost sensitive loss}. When training our ML classifier, its threshold was chosen in alignment with the Neyman-Pearson criterion, so we set $\lambda$ based on the model's threshold $t$.

\subsection{Baselines}

When searching for possible L2D baselines, we found no current method able to take individual capacity constraints into account. Therefore, we provide three baselines. 

\textbf{Rejection Learning (ReL)} In this implementation we use the model score as a measure of model confidence. To apply rejection learning batch-wise, within our capacity constraints, we first order the cases within the batch by descending order of model score. The top 5\% cases are automatically predicted positive (declined). Then, the following cases are randomly assigned to experts within our team until their capacity constraints are met. All left over cases, with the lowest model score, are classified negatively (accepted).

\textbf{Human Expertise Aware Rejection Learning} In this version of \textit{rejection learning}, instead of randomly assigning the rejected cases throughout our expert team, we attempt to model each individual's behavior, in order to optimize assignments. To do so, we train a LightGBM model on the instance features and the \textit{expert\_id} to predict if the expert's prediction was a false positive (FP), false negative (FN), true positive (TP), or true negative (TN). For each instance, we then have a prediction for the probability that the expert will make either a FP, or a FN mistake, $\hat{\mathbb{P}}(\mbox{FP})$ and $\hat{\mathbb{P}}(\mbox{FN})$, respectively. We then calculate the predicted loss associated with deferring instance $\mathbf{x}_i$ to expert \textit{e}:
\begin{equation}
\label{eq: Cost sensitive formulation}
    L(\mathbf{X}_i, e) = \lambda \hat{\mathbb{P}}(\mbox{FP}) + \hat{\mathbb{P}}(\mbox{FN}) \;,
\end{equation}

We present two versions of this algorithm:
\begin{itemize}
    \item \textbf{Greedy ($\mbox{ReL}_{\mbox{\scriptsize greedy}}$)}: The algorithm moves through the batch case by case, assigning each case to the expert with lowest predicted loss. Should an assignment violate capacity constraints, the algorithm tries to assign to the expert with second lowest loss, and so on. This is done until the experts' capacity constraints are met.
    \item \textbf{Integer Linear Programming ($\mbox{ReL}_{\mbox{\scriptsize linear}}$)}: In this method, we minimize the loss over an entire batch by solving a linear programming problem subject to our capacity constraints, in order to find the optimal assignment over the entire batch.
\end{itemize}

For the context of fairness,
we want to guarantee that the probability of wrongly declining a legitimate application is
independent of the sensitive attribute value. Hence we measure the ratio between
FPRs in each age group, i.e., predictive equality (PE) \cite{corbett2017algorithmic}. The ratio is calculated by dividing the FPR of the group with lowest observed FPR by the FPR of the group with the highest FPR.

\subsection{Results}

In Table ~\ref{table: baseline results} we show results for the discussed L2D baselines as well as a ``Model only'' system. The loss function is calculated according to Equation \ref{eq: cost sensitive loss}, with N(FP) and N(FN) counted over the test set. We can see how results for each of our L2D baselines vary with the generated human-AI collaboration environment (Scenario Properties) across the rows for our performance metric (Cost sensitive loss) and our fairness metric (Predictive Equality). 
We observe that, throughout all the considered scenarios, rejection learning performs best, despite our attempts to model human behavior. This may be due to the low volume of FNs and TPs in the training data, which may lead to poor probability estimates and ranking of the expert's probability of error for a given instance. In section D of the Appendix, Table 6 shows that our methods mostly mitigate FP errors, resulting in a lower FPR, but negatively impacting the recall as well. The mitigation of False Positives also leads to higher predictive equality, showing that our human expertise model was able to learn that experts tend to make more FP mistakes on older clients' applications.
A drastic variation can be seen in the results for the $\mbox{ReL}_{\mbox{\scriptsize linear}}$ method. While it seems that expert absence has no significant effect on the loss for scenarios with a deferral rate of 20\%, in the cases with 50\% deferral rate, it seems that introducing expert absence significantly increases the loss.
This illustrates the importance of testing the system under a wide variety of conditions.

In Table ~\ref{table: baseline results pool}, we introduce variation in the pool of available experts. Here we can see that $\mbox{ReL}_{\mbox{\scriptsize linear}}$ outperforms ReL when the expert pool contains only agreeing, sparse or unfair experts. This may be due to the fact that these experts are simpler to model, as they have a clear dominating feature, or simpler feature dependencies. This illustrates the importance of considering variable complexity of human behavior when evaluating HAIC systems.

\section{Conclusions and Future Work}

In this paper, we introduced the FiFAR Dataset. To illustrate its use, we provide three L2D baselines tested under 300 different scenarios. Our dataset enables comprehensive benchmarking of L2D algorithms, subject to real world constraints and scenarios.
The main limitation of our work is that our baselines do not include any established methods in the L2D literature, as these do not currently consider the existence of capacity constraints.

We emphasize that our synthetic experts can not replace, in any way, humans involved in HAIC systems, as it would be necessary to gather real expert data in order to train the system to be used in a real-world application. 
It can be argued that, by using these synthetic experts for research purposes, we are affecting the livelihood of large-scale annotation workers (i.e. MTurk), which are often used by researchers. However, in some use cases, where domain-specific expertise is needed, such annotation services may not be adequate, and it may be impossible/unfeasible to obtain real human expert data. In cases where human predictions are accessible and pertinent to the use case, researchers should prefer these over synthetic expert predictions, as they constitute real human behavior. 
For these reasons, we believe that our work motivates the use of more complex synthetic expert data when real human predictions are unavailable, without posing a threat to current existing annotation services. It is also important to emphasize that our synthetic experts may amplify biases, due to the fact that they establish a monotonic relationship between each feature and the probabilities of error. It is possible that by increasing the weight of a feature that is correlated with the protected attribute, an expert with a positive weight for said feature in the false positive probability would exhibit a higher bias against said protected group. As such, biases could be amplified by our synthetic agents, and a careful analysis of the final predictive equality of each expert is encouraged.

\newpage

\newpage

\bibliographystyle{ACM-Reference-Format}
\bibliography{sample-base}
  
\end{document}